\documentclass[letterpaper, 10 pt, journal, twoside]{ieeetran}

\usepackage{multirow}
\usepackage{booktabs}
\usepackage{tabularx}
\usepackage{amsmath}
\usepackage{amsfonts,amssymb}
\usepackage{graphicx} % for pdf, bitmapped graphics files
\usepackage{subfigure}
\usepackage[ruled,linesnumbered]{algorithm2e}
\graphicspath{{./}}
\usepackage{cite}
\usepackage{color}
\usepackage[breaklinks, colorlinks, linkcolor=black, citecolor=black, urlcolor=black]{hyperref}
\usepackage{verbatim}
\usepackage{url}
\usepackage{flushend}
\begin{document}
	
\title{	Open-Set Object Detection Using Classification-free Object Proposal and Instance-level Contrastive Learning }
\author{Zhongxiang Zhou$^{1}$, Yifei Yang$^{1}$, Yue Wang$^{1}$, Rong Xiong$^{1}$% <-this % stops a space
\thanks{Manuscript received: November 22, 2022; Accepted January 16, 2023.}%Use only for final RAL version
\thanks{This paper was recommended for publication by Editor Cesar Cadena upon evaluation of the Associate Editor and Reviewers’ comments. This work was supported by the National Nature Science Foundation of China (62173293) and Zhejiang Provincial Natural Science Foundation of China (LD22E050007).}%Use only for final RAL version
\thanks{$^{1}$Zhongxiang Zhou, Yifei Yang, Yue Wang and Rong Xiong are with the State Key Laboratory of Industrial Control Technology and Institute of Cyber-Systems and Control, Zhejiang University, Zhejiang, China. Yue Wang is the corresponding author {\tt\small ywang24@zju.edu.cn}}%
\thanks{Digital Object Identifier (DOI): see top of this page.}
}
% Use only for final RAL version.

\markboth{IEEE Robotics and Automation Letters. Preprint Version. Accepted January, 2023}
{Zhou \MakeLowercase{\textit{et al.}}: Open-Set Object Detection Using Classification-free Object Proposal and Instance-level Contrastive Learning} 

\maketitle

%%%%%%%%%%%%%%%%%%%%%%%%%%%%%%%%%%%%%%%%%%%%%%%%%%%%%%%%%%%%%%%%%%%%%%%%%%%%%%%%
\begin{abstract}

Detecting both known and unknown objects is a fundamental skill for robot manipulation in unstructured environments. Open-set object detection (OSOD) is a promising direction to handle the problem consisting of two subtasks: objects and background separation, and open-set object classification. In this paper, we present Openset RCNN to address the challenging OSOD. To disambiguate unknown objects and background in the first subtask, we propose to use classification-free region proposal network (CF-RPN) which estimates the objectness score of each region purely using cues from object's location and shape preventing overfitting to the training categories. To identify unknown objects in the second subtask, we propose to represent them using the complementary region of known categories in a latent space which is accomplished by a prototype learning network (PLN). PLN performs instance-level contrastive learning to encode proposals to a latent space and builds a compact region centering with a prototype for each known category. Further, we note that the detection performance of unknown objects can not be unbiasedly evaluated on the situation that commonly used object detection datasets are not fully annotated. Thus, a new benchmark is introduced by reorganizing GraspNet-1billion, a robotic grasp pose detection dataset with complete annotation. Extensive experiments demonstrate the merits of our method. We finally show that our Openset RCNN can endow the robot with an open-set perception ability to support robotic rearrangement tasks in cluttered environments. More details can be found in \href{https://sites.google.com/view/openset-rcnn/}{https://sites.google.com/view/openset-rcnn/}

\end{abstract}

\begin{IEEEkeywords}
Perception for Grasping and Manipulation; Object Detection, Segmentation and Categorization
\end{IEEEkeywords}

\IEEEpeerreviewmaketitle
%%%%%%%%%%%%%%%%%%%%%%%%%%%%%%%%%%%%%%%%%%%%%%%%%%%%%%%%%%%%%%%%%%%%%%%%%%%%%%%%
\section{Introduction}

\IEEEPARstart{W}{hen} a robot executes manipulation tasks in unstructured environments, it will encounter unknown objects together with known objects. The task may fail if unknown objects are not being correctly perceived. Although significant achievements have been made in object detection  \cite{ren2015faster,redmon2016you,carion2020end}, most object detectors assume that only a subset of training categories will appear in testing phase. Since object categories in real-world are infinite, close-set object detectors are not practical for the robot to work in unstructured environments.

\begin{figure}[t]
	\centering
	\setlength{\abovecaptionskip}{0.cm}
	\includegraphics[scale=0.305]{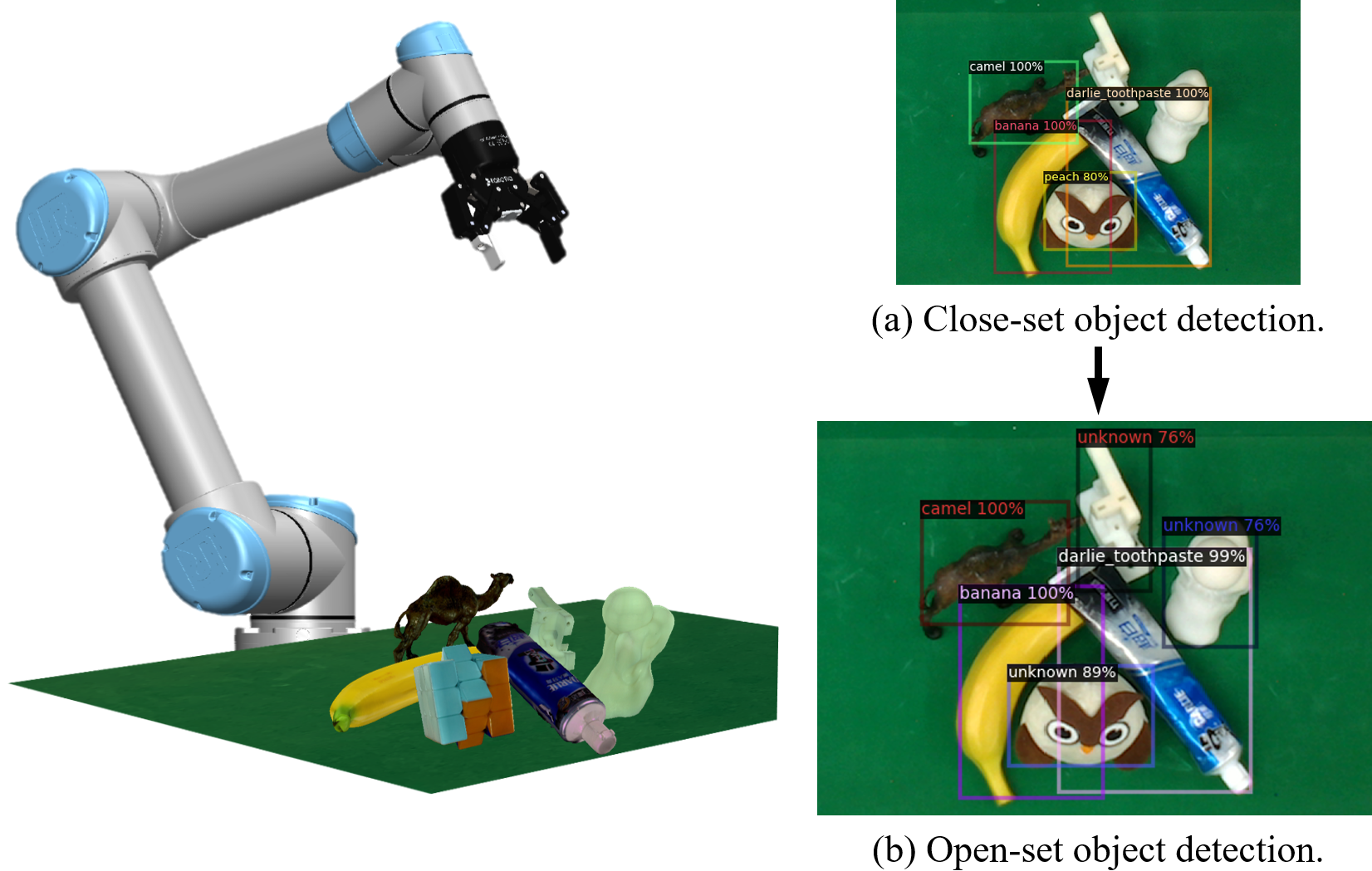}
	\caption{(a) Close-set object detection assumes that object categories appearing in the testing phase are contained by training categories. When encountering unknown objects, a close-set object detector tends to simply ignore them or misclassify unknown objects to known categories. (b) Open-set object detection also trains a model on \textit{close-set} datasets, but it is asked to detect all known objects and unknown objects in \textit{open-set} conditions.}
	\vspace{-0.3cm}
	\label{fig:fig1}
\end{figure}

To endow the robot's ability to detect both known and unknown objects, a potential solution is open-set object detection (OSOD), where the detector trained on \textit{close-set} datasets is asked to detect all known objects and identify unknown objects in \textit{open-set} conditions \cite{han2022opendet}. Apparently, OSOD is composed of two subtasks: 1) objects and background separation, 2) open-set object classification. An illustration showing the difference between close-set and open-set object detection for robotic manipulation is presented in Fig. \ref{fig:fig1}. %Fig. \ref{fig:fig1} illustrates the difference between close-set and open-set object detection for robotic manipulation. %An intuitive illustration showing the difference between close-set and open-set object detection for robotic manipulation is presented in Fig. \ref{fig:fig1}.

Lately, several works attempted to solve the challenging OSOD. Dhamija et al. \cite{dhamija2020overlooked} first formalized OSOD task and concluded that close-set object detectors suffer from performance degradation in the open-set conditions. Miller et al. \cite{miller2018dropout, miller2021uncertainty} studied the utility of Dropout Sampling \cite{gal2016dropout} and Gaussian Mixture Models for estimating uncertainty which is used to identify open-set errors. Joseph et al. \cite{joseph2021towards} proposed an energy-based unknown identifier by modeling the energy values of known and unknown categories. Han et al. \cite{han2022opendet} presented an unknown probability learner which is used to directly predict the known and unknown probabilities of proposals. However, prior works focus on the second subtask of OSOD improving the open-set classification ability. And unknown objects are treated in the same way as one of the known classes which is characterized using a point or region in latent space. We note that unknown objects may contain more than one category, representing all unknown objects using a single latent point or region is inaccurate.

As for how to separate objects and background, previous methods rely on classification based Region Proposal Network (RPN) \cite{ren2015faster}. Recently, Kim et al. \cite{kim2022learning} noticed that RPN tends to overfit to the training categories if using the binary classifier for close-set training. Because proposals for unknown objects would be regarded as negatives by the classification loss, impeding OSOD performance. Moreover, commonly used datasets such as PASCAL VOC \cite{everingham2010pascal} and COCO \cite{lin2014microsoft} do not exhaustively label all objects, detections on unannotated objects will be regarded as false positives making the evaluation of unknown objects biased.

In this paper, we present Openset RCNN with classification-free RPN (CF-RPN) and prototype learning network (PLN) to address the challenging OSOD. CF-RPN learns object proposals using cues from object's location and shape without classification to enhance generalization ability for unknown object proposals. PLN performs instance-level contrastive learning to encode proposals in a latent space and learning a discriminative representation called prototype for each known category. Benefiting from intra-class compactness and inter-class separation in the learnt latent space, known and unknown objects can be identified using distances between encoded proposals and prototypes. In this way, the complementary region of known categories in latent space can be used to represent unknown objects covering more unknown categories. Followed by a softmax classifier, known proposals are further classified into known categories.

To validate the effectiveness of our method, we first take VOC for close-set training and conduct open-set testing using both VOC and COCO which is the same as \cite{han2022opendet}. Differently, we do not adopt Average Precision of unknown as the evaluation metric since objects in VOC and COCO are not fully annotated. Furthermore, GraspNet-1billion \cite{fang2020graspnet}, a robotic grasp pose detection dataset with complete annotation, is reorganized to form a new benchmark considering several open-set settings for fair comparison. Experimental results demonstrate that our method improves OSOD performance. Finally, we apply the proposed method to five robotic rearrangement tasks to manipulate known and unknown objects differently. Our contributions can be summarized as follows:
\begin{itemize}
	\item A novel method, Openset RCNN, is proposed to enhance generalization and identification ability for unknown objects with the help of CF-RPN and instance-level contrastive learning based PLN. 
	\item We introduce a new OSOD benchmark by reorganizing a fully annotated dataset called GraspNet-1billion that can be used for an unbiased evaluation for OSOD and is more suitable for robot manipulation.
	\item Extensive experiments demonstrate the merits of our method. And we show that our method can support robotic rearrangement tasks in cluttered environments. %be applied to the robotic rearrangement tasks in cluttered environments supporting the robot to manipulate known and unknwon objects differently.
\end{itemize}

\section{Related Work}

\subsection{Open Set Recognition}

Scheirer et al. \cite{scheirer2012toward} first formalized OSR as an open risk minimization problem and developed an open-set classifier using linear SVMs. Their pioneering work inspired follow-up SVM-based approaches \cite{scheirer2014probability, jain2014multi, bendale2015towards}. Recently, deep learning-based OSR methods have aroused extensive attention. Bendale et al. \cite{bendale2016towards} introduced the first deep learning approach for OSR by replacing the SoftMax function with OpenMax function. Neal et al. \cite{neal2018open} utilized GAN to synthesize unseen-class images and trained an open-set classifier with the generated images augmented training dataset. OpenGAN \cite{kong2021opengan} found that it was more efficient to generate open-set features rather than realistic images, then an adversarially trained discriminator was used to classify unknown examples. Other approaches include auto-encoder-based methods which use reconstruction errors as indicators of unknown identification \cite{sun2020conditional}, and prototype-based methods which identify unknowns by measuring the distances between image features and learned prototypes \cite{yang2020convolutional}. In addition, Vaze et al. \cite{vaze2022openset} investigated that good close-set classifier benefits OSR. However, OSR can only recognize objects at the image level which limits its applications in robotic manipulation tasks.

\subsection{Open Set Object Detection}

Dhamija et al. \cite{dhamija2020overlooked} first migrated the open-set setting to the object detection field. They noticed that objects from unknown classes were often incorrectly detected as known objects. In the context that uncertainty may indicate open-set errors, Miller et al. \cite{miller2018dropout, miller2021uncertainty} studied the utility of Dropout Sampling \cite{gal2016dropout} and Gaussian Mixture Models to analyze the detector’s epistemic uncertainty and predictions with high uncertainty were regarded as unknown objects. Several works are proposed by introducing unknown labels during training. Joseph et al. \cite{joseph2021towards} introduced an energy-based unknown identifier by modeling the energy values of known and unknown samples. However, annotations of unknown objects are needed by their methods \cite{joseph2021towards, singhorder} breaking the setting of OSOD. Based on a transformer-based object detector, OW-DETR \cite{gupta2022ow} adopted the magnitude of the attention feature map as the objectness scores of proposals and top-k selection was performed for obtaining pseudo-unknowns. The pseudo-labeled unknowns along with the ground-truth known objects are employed as foreground objects during training. UC-OWOD \cite{wu2022uc} also used the same strategy to select unknown samples from unannotated areas. Han et al. \cite{han2022opendet} presented an unknown probability learner augmenting the K-way classifier with the K+1-way classifier where K+1 denoted the unknown class. In order to reserve the probability for unknown classes during close-set training, they considered that the probability of unknown class should be the largest if the logit of ground-truth class is removed. In summary, previous OSOD methods treat unknown objects in the same way as one of the known classes which is characterized using a point or region in latent space. Since unknown objects may contain more than one category, this characterization is inaccurate. In contrast, our method represents unknown objects using the complementary region of known classes covering more unknown categories.

\begin{figure*}[th]
	\centering
	\setlength{\abovecaptionskip}{0.cm}
	\includegraphics[scale=0.57]{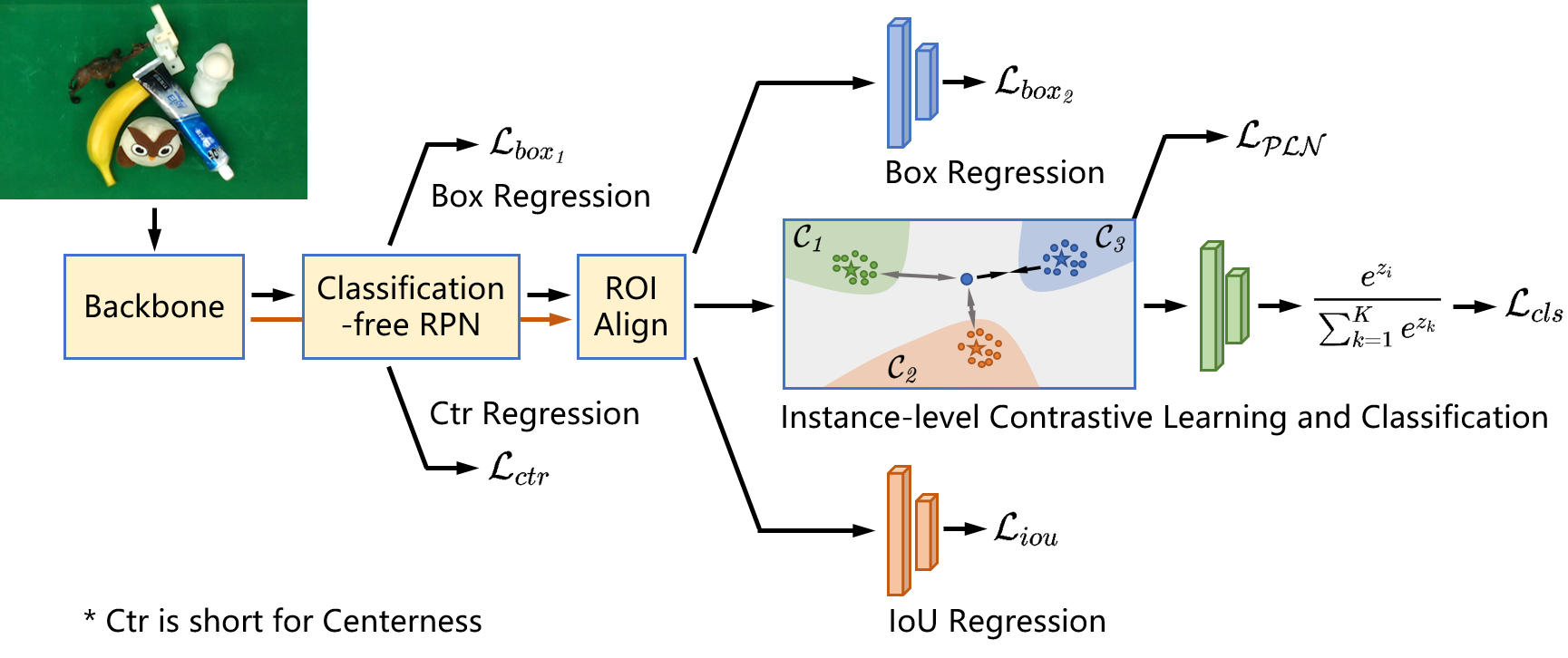}
	\caption{Overview of the proposed open-set object detection method. Openset RCNN is endowed with classification-free RPN (CF-RPN) and prototype learning network (PLN) perfroming instance-level contrastive learning. CF-RPN learns to detect object proposals based on cues from object's location and shape. PLN constructs a latent space which can be used to identify known and unknown objects using distances between encoded proposals and prototypes. Followed by a softmax classifier, known proposals are further classified into known categories.}
	\label{fig:fig2}
	\vspace{-0.1cm}
\end{figure*}

\subsection{Contrastive Learning}
Self-supervised representation learning has recently achieved state of the art performance with the help of contrastive learning \cite{he2020momentum, tian2020makes, qian2021spatiotemporal}. To effectively leverage label information, supervised contrastive learning \cite{khosla2020supervised} has been proposed and applied to other tasks, such as few-shot object detection \cite{sun2021fsce}, semantic segmentation \cite{zhao2021contrastive} and zero-shot learning \cite{han2021contrastive}. Our method applies instance-level supervised contrastive learning to encode proposals in a latent space and build a compact region centering with a prototype for each known category.

\section{Problem Formalization}

We formally describe OSOD based on previous researches \cite{dhamija2020overlooked, joseph2021towards, han2022opendet}. We denote two datasets $D_{tr} = \{(x,y), x \in X_{tr}, y \in Y_{tr}\}$ and $D_{te} = \{(x',y'), x' \in X_{te}, y' \in Y_{te}\}$, where $x$ and $x'$ are input images with $y$ and $y'$ representing corresponding annotations including category label and coordinates of bounding box. An open-set object detector $\mathcal{M}$ is trained using $D_{tr}$ with image set $X_{tr}$ and annotation set $Y_{tr}$ containing $K$ known categories $C_K = \{c_1,...,c_k\}$. Then $\mathcal{M}$ is tested using $D_{te}$ with image set $X_{te}$ and annotation set $Y_{te}$ containing $K$ known categories as well as other unknown objects $c_u \notin C_K$.  And $\mathcal{M}$ is supposed to detect all known objects belonging to $C_K$ and unknown objects $c_u$ in $D_{te}$.

In general, OSOD is composed of two subtasks: 1) objects and background separation, 2) open-set object classification. And two unique challenges are held by OSOD. Since $D_{tr}$ may not be fully annotated, images in $X_{tr}$ may contain annotated known objects and unannotated unknown objects. Accordingly, overfitting issue must be well addressed in OSOD. At the same time, complete annotation must be given in $D_{te}$ in order to evaluate OSOD performance, especially on unknown objects. 

\section{Methodology} 

We present Openset RCNN with classification-free RPN (CF-RPN) and instance-level contrastive learning based prototype learning network (PLN). The framework is shown in Fig. \ref{fig:fig2}. ResNet-50 \cite{he2016deep} with Feature Pyramid Network (FPN) \cite{lin2017feature} is used as the backbone network to extract feature maps of the input image. Classification-free RPN learns to detect object proposals based on cues from object's location and shape. Then, instance-level contrastive learning is performed to encode proposals in a latent space and learn a discriminative representation called prototype for each known category. After that, known and unknown objects can be identified using distances between encoded proposals and prototypes. Followed by a softmax classifier, known proposals are further classified into known categories.

\subsection{Classification-free Region Proposal Network}
\begin{figure} 
	\centering 
	\setlength{\abovecaptionskip}{0.cm}
	\subfigure[]{ 
		\label{fig:fig3:a} %% label for first subfigure 
		\includegraphics[scale=0.5]{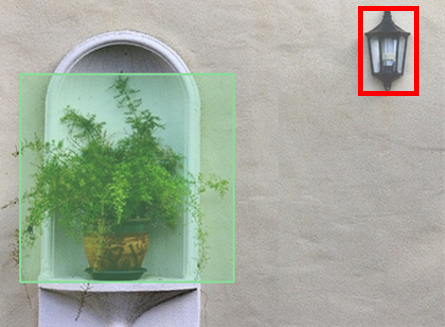}}
	\hspace{-0.1cm}
	\subfigure[]{ 
		\label{fig:fig3:b} %% label for second subfigure 
		\includegraphics[scale=0.5]{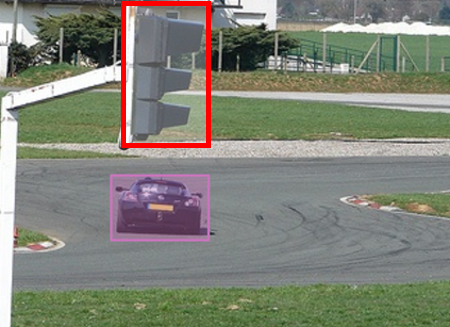}}
	
	\vspace{-0.2cm}
	\subfigure[]{ 
		\label{fig:fig3:c} %% label for second subfigure 
		\includegraphics[scale=0.5]{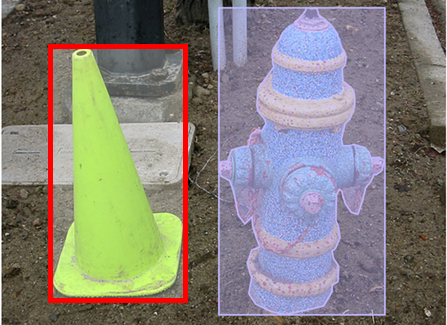}}
	\hspace{-0.1cm}
	\subfigure[]{ 
		\label{fig:fig3:d} %% label for second subfigure 
		\includegraphics[scale=0.5]{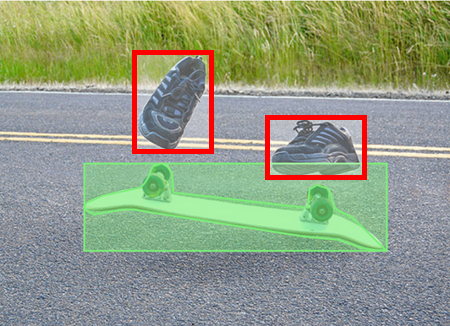}}
	\caption{Commonly used object detection datasets are not fully annotated, (a) PASCAL VOC trainval2007, (b) PASCAL VOC test2007, (c) COCO train2017 and (d) COCO val2017. Red boxes are objects without annotations and other colored boxes are annotated objects. } 
	\vspace{-0.5cm}
	\label{fig:fig3} %% label for entire figure 
\end{figure}

To extract foreground objects in the image, classification based RPN \cite{ren2015faster} is popularly used. But classification based RPN often struggles to detect novel objects in open-set conditions because of overfitting to training categories. The overfitting issue is caused by two parts. On one hand, commonly used object detection datasets such as PASCAL VOC and COCO do not exhaustively label all objects, as shown in Fig. \ref{fig:fig3}. On the other hand, binary classification loss is optimized with training categories as foreground and other unannotated objects as background.

Inspired by \cite{kim2022learning}, we use CF-RPN in our open-set object detector to enhance generalization ability for unknown object proposals. Without supervision from category label, CF-RPN learns to detect objects using cues from object location and shape called localization quality. Specifically, binary classifier in standard RPN is replaced with centerness \cite{tian2019fcos} regression. And $ltrb$ bounding box regression is used to generate initial proposals instead of delta $xywh$ regression according to the definition of centercess. Top-scoring proposals are taken to perform RoIAlign \cite{he2017mask}, then fed to bounding box refinement head and intersaction-over-union (IoU) regression head. Both centerness score $c$ and IoU score $b$ are adopted to measure the quality of object proposal. Following \cite{kim2022learning}, objectness score is computed as a geometric mean of these two values $s = \sqrt{c \cdot b}$.

During training, a proposal is considered as a positive sample if it has IoU greater than $T_{pos}$ with the corresponding ground-truth box and a negative sample if IoU less than $T_{neg}$. Then, $N_s$ proposals are randomly sampled with positive fraction $P_{pos}$ for loss computation. Our CF-RPN loss is formulated as:
\begin{equation}
\mathcal{L_{CF-RPN}} = \lambda_1\mathcal{L}\mathit{_{ctr}} + \lambda_2\mathcal{L}\mathit{_{box_1}} + \lambda_3\mathcal{L}\mathit{_{iou}} + \lambda_4\mathcal{L}\mathit{_{box_2}}
\label{equ:equ1}
\end{equation}
where $\mathcal{L}\mathit{_{ctr}}$, $\mathcal{L}\mathit{_{box_1}}$, $\mathcal{L}\mathit{_{iou}}$ and $\mathcal{L}\mathit{_{box_2}}$ are smooth L1 losses for centerness regression, $ltrb$ bounding box regression, IoU regression and delta $xywh$ bounding box regression. And $\lambda_1$, $\lambda_2$, $\lambda_3$, and $\lambda_4$ are weighting coefficients. The first two losses are used for initial proposal generation and the last two losses are used for proposal refinement.

\subsection{Prototype Learning Network}
To identify known and unknown objects from proposals generated by CF-RPN, we use PLN to construct a latent space and learn a discriminative representation called prototype for each known category. Then, known and unknown objects can be identified using distances between encoded proposals and prototypes.

Our PLN uses the proposal feature $\mathbf{f}_i \in \mathbb{R}^{d_f}$ after RoIAlign as input. Typically, $d_f=1024$ which is the same as Faster RCNN \cite{ren2015faster}. Then, we use an encoder to map $\mathbf{f}_i$ to a low-dimensional embedding vector $\mathbf{z}_i \in \mathbb{R}^{d_z}$ ($d_z=256$ by default) in the latent space. In detail, the encoder is a single fully connected layer with nonlinearity. Subsequently, we measure distances between proposal embedding $\mathbf{z}_i$ and learnable prototypes $\mathbf{P} = \{\mathbf{P}_1,...,\mathbf{P}_K\}$ of known categories by computing the cosine similarity. Different from \cite{joseph2021towards}, we use the weights of a fully connected layer as prototypes with less memory costs while they need to maintain a fixed length queue for each known category to store the embedding vectors and prototypes are computed by averaging all vectors in the queue. Next, intra-class compactness and inter-class separation are encouraged by optimizing the following double-margin contrastive loss:
\begin{equation}
\begin{split}
\mathcal{L_{PLN}} = &y_{ij}\max(D_{ij} - m_p, 0)\\
					&+ \mathop{max}\limits_j\left[(1-y_{ij})\max(m_n-D_{ij}, 0)\right]
\end{split}
\label{equ:equ2}
\end{equation}

where $D_{ij}=1-\frac{\mathbf{z}_i \cdot \mathbf{P}_j}{\left\Vert\mathbf{z}_i\right\Vert \left\Vert\mathbf{P}_j\right\Vert}$ is the cosine distance between proposal embedding $\mathbf{z}_i$ and prototype $\mathbf{P}_j \in \mathbf{P} $, $y_{ij}$ is an indicating variable with value 1 if $\mathbf{z}_i$ and $\mathbf{P}_j$ are from the same category and 0 otherwise. As for $m_p$ and $m_n$, they are thresholds for positive pairs from the same category and negative pairs from different categories. Contrastive loss $\mathcal{L_{PLN}} = 0$ under the condition that distances between proposal embeddings and prototype of the same category are less than $m_p$ and distances between proposal embeddings and prototypes of different categories are larger than $m_n$. Note that we optimize the latent space using proposals having IoU $> T_{iou}$ in a mini-batch where $T_{iou}$ is an IoU threshold between proposals and ground-truth boxes to ensure the proposals contain relevant semantics.

Although the latent space can be used to measure the distances between a proposal embedding and each prototype, the category of this proposal can be determined by the prototype with minimum distance. We find that the close-set classification performance of the latent space is not as well as the traditional softmax classifier. So after identifying known and unknown objects using the latent space constructed by PLN, we apply a softmax classifier on all known objects to classify them into each known category. And the 256-dim proposal embedding $\mathbf{z}_i$ is remapped to 1024-dim to restore the detail information benefiting classification. The loss function $\mathcal{L}_{cls}$ used for optimization is a cross entropy loss which is the same as Faster RCNN.

\subsection{Overall Optimization and Inference}

In total, the network is trained by the following multi-task loss:
\begin{equation}
\mathcal{L} = \alpha\mathcal{L_{CF-RPN}} + \beta\mathcal{L_{PLN}} + \gamma\mathcal{L}_{cls}
\label{equ:equ3}
\end{equation}

which is a conbination of $\mathcal{L_{CF-RPN}}$, $\mathcal{L_{PLN}}$ and $\mathcal{L}_{cls}$ with weighting coefficients $\alpha$, $\beta$ and $\gamma$. To train CF-RPN, we set $N_s=256$, $T_{pos}=0.3$, $T_{neg}=0.1$ and $P_{pos}=1.0$ for optimizing centerness regression loss $\mathit{L_{ctr}}$, set $N_s=256$, $T_{pos}=0.7$, $T_{neg}=0.3$ and $P_{pos}=0.5$ for optimizing $ltrb$ bounding box regression loss $\mathit{L_{box_1}}$, and set $N_s=512$, $T_{pos}=0.5$, $T_{neg}=0.5$ and $P_{pos}=0.25$ for optimizing IoU regression loss and delta $xywh$ regression loss. We set $m_p=0.1$, $m_n=0.9$, $T_{iou}=0.5$, $\alpha=1$, $\beta=0.5$, $\gamma=0.9$ and $\lambda_{1,2,3,4}=0.5$ in order to optimize network on OpenDet benchmark, and set $m_p=0.05$, $m_n=0.95$, $T_{iou}=0.5$, $\alpha=1$, $\beta=2$, $\gamma=1.5$ and $\lambda_{1,2,3,4}=\{1,10,1,2\}$ for optimizing on GraspNet-1billion dataset. Ablation studies on hyper-parameters are available in Appendix-C \cite{ours2022arxiv}.

During training, top $2000$ centerness scoring proposals without non-maximum suppression (NMS) are taken for proposal refinement and the subsequent process. During inference, top $1000$ centerness scoring proposals are sent to NMS at $0.7$, RoIAlign and proposal refinement. Then, objectness scores $s$ are calculated for each proposal following a filter to remove proposals with $s<0.05$. Next, proposals are encoded to the latent space to compute distances between each prototype and unknown objects are identified if the minimum distance among all prototypes is larger than a predefined threshold $T_u$. Subsequently, known objects are sent to softmax classifier to classify them into each category. Note that after NMS for known and unknown objects separately, top $50$ objectness scoring known and top 50 objectness scoring unknown objects are selected as the final detections.

\begin{table*}[ht]
	\centering
	\setlength{\abovecaptionskip}{0.cm}
	\caption{Comparisons with other methods on VOC and VOC-COCO-T1.}
	\renewcommand\arraystretch{1.5}
	\label{table:opendet_t_1}
	\resizebox{\textwidth}{!}{
	\begin{tabular}{l|l|llll|llll|llll}
		\hline
		\multicolumn{1}{l|}{\multirow{2}{*}{Method}} & \multicolumn{1}{c|}{VOC}   & \multicolumn{4}{c|}{VOC-COCO-20}                                                                               & \multicolumn{4}{c|}{VOC-COCO-40}                                                                               & \multicolumn{4}{c}{VOC-COCO-80}                                                                               \\ \cline{2-14} 
		\multicolumn{1}{c|}{}                        & \multicolumn{1}{c|}{mAP$_\mathcal{K}{_\uparrow}$}  & \multicolumn{1}{c}{WI$_\downarrow$}    & \multicolumn{1}{c}{AOSE$_\downarrow$}  & \multicolumn{1}{c}{mAP$_\mathcal{K}{_\uparrow}$}  & \multicolumn{1}{c|}{R$_\mathcal{U}{_\uparrow}$}   & \multicolumn{1}{c}{WI$_\downarrow$}    & \multicolumn{1}{c}{AOSE$_\downarrow$}  & \multicolumn{1}{c}{mAP$_\mathcal{K}{_\uparrow}$}  & \multicolumn{1}{c|}{R$_\mathcal{U}{_\uparrow}$}   & \multicolumn{1}{c}{WI$_\downarrow$}    & \multicolumn{1}{c}{AOSE$_\downarrow$}  & \multicolumn{1}{c}{mAP$_\mathcal{K}{_\uparrow}$}  & \multicolumn{1}{c}{R$_\mathcal{U}{_\uparrow}$}   \\ \hline
		FR-CNN \cite{ren2015faster}                           & \multicolumn{1}{c|}{82.68} & \multicolumn{1}{c}{21.43} & \multicolumn{1}{c}{15058} & \multicolumn{1}{c}{58.22} & \multicolumn{1}{c|}{0.00} & \multicolumn{1}{c}{25.92} & \multicolumn{1}{c}{23070} & \multicolumn{1}{c}{55.01} & \multicolumn{1}{c|}{0.00} & \multicolumn{1}{c}{20.25} & \multicolumn{1}{c}{23216} & \multicolumn{1}{c}{55.44} & \multicolumn{1}{c}{0.00} \\
		PROSER \cite{zhou2021learning}                        & \multicolumn{1}{c|}{81.84} & \multicolumn{1}{c}{21.18} & \multicolumn{1}{c}{13993} & \multicolumn{1}{c}{56.97} & \multicolumn{1}{c|}{37.85} & \multicolumn{1}{c}{26.13} & \multicolumn{1}{c}{21676} & \multicolumn{1}{c}{53.86} & \multicolumn{1}{c|}{25.39} & \multicolumn{1}{c}{20.55} & \multicolumn{1}{c}{22499} & \multicolumn{1}{c}{54.60} & \multicolumn{1}{c}{12.75} \\
		DS \cite{miller2018dropout}                           & \multicolumn{1}{c|}{82.51} & \multicolumn{1}{c}{17.30} & \multicolumn{1}{c}{13031} & \multicolumn{1}{c}{58.39} & \multicolumn{1}{c|}{19.67} & \multicolumn{1}{c}{20.93} & \multicolumn{1}{c}{19920} & \multicolumn{1}{c}{55.41} & \multicolumn{1}{c|}{13.04} & \multicolumn{1}{c}{16.76} & \multicolumn{1}{c}{21749} & \multicolumn{1}{c}{56.11} & \multicolumn{1}{c}{5.38} \\
		OpenDet \cite{han2022opendet}                        & \multicolumn{1}{c|}{82.09} & \multicolumn{1}{c}{13.71} & \multicolumn{1}{c}{12080} & \multicolumn{1}{c}{58.08} & \multicolumn{1}{c|}{36.52} & \multicolumn{1}{c}{16.56} & \multicolumn{1}{c}{18116} & \multicolumn{1}{c}{54.90} & \multicolumn{1}{c|}{24.87} & \multicolumn{1}{c}{13.38} & \multicolumn{1}{c}{20790} & \multicolumn{1}{c}{55.88} & \multicolumn{1}{c}{12.79} \\ \hline
		Openset RCNN                                & \multicolumn{1}{c|}{82.58} & \multicolumn{1}{c}{\textbf{11.77}} & \multicolumn{1}{c}{\textbf{10452}}  & \multicolumn{1}{c}{\textbf{58.96}} & \multicolumn{1}{c|}{\textbf{53.65}} & \multicolumn{1}{c}{\textbf{14.70}} & \multicolumn{1}{c}{\textbf{16065}} & \multicolumn{1}{c}{\textbf{55.65}} & \multicolumn{1}{c|}{\textbf{43.83}} & \multicolumn{1}{c}{\textbf{12.36}} & \multicolumn{1}{c}{\textbf{18778}} & \multicolumn{1}{c}{\textbf{56.48}} & \multicolumn{1}{c}{\textbf{31.94}} \\ \hline
	\end{tabular} }
\end{table*}

\begin{table*}[ht]
	\centering
	\setlength{\abovecaptionskip}{0.cm}
	\caption{Comparisons with other methods on VOC-COCO-T2.}
	\renewcommand\arraystretch{1.5}
	\label{table:opendet_t_2}
	\setlength{\tabcolsep}{2.6mm}
	\resizebox{\textwidth}{!}{
	\begin{tabular}{l|llll|llll|llll}
		\hline
		\multicolumn{1}{l|}{\multirow{2}{*}{Method}}   & \multicolumn{4}{c|}{VOC-COCO-2500}                                                                               & \multicolumn{4}{c|}{VOC-COCO-5000}                                                                               & \multicolumn{4}{c}{VOC-COCO-20000}                                                                               \\ \cline{2-13} 
		\multicolumn{1}{c|}{}                        & \multicolumn{1}{c}{WI$_\downarrow$}    & \multicolumn{1}{c}{AOSE$_\downarrow$}  & \multicolumn{1}{c}{mAP$_\mathcal{K}{_\uparrow}$}  & \multicolumn{1}{c|}{R$_\mathcal{U}{_\uparrow}$}   & \multicolumn{1}{c}{WI$_\downarrow$}    & \multicolumn{1}{c}{AOSE$_\downarrow$}  & \multicolumn{1}{c}{mAP$_\mathcal{K}{_\uparrow}$}  & \multicolumn{1}{c|}{R$_\mathcal{U}{_\uparrow}$}   & \multicolumn{1}{c}{WI$_\downarrow$}    & \multicolumn{1}{c}{AOSE$_\downarrow$}  & \multicolumn{1}{c}{mAP$_\mathcal{K}{_\uparrow}$}  & \multicolumn{1}{c}{R$_\mathcal{U}{_\uparrow}$}   \\ \hline
		FR-CNN \cite{ren2015faster}                           & \multicolumn{1}{c}{8.93}  & \multicolumn{1}{c}{5795} & \multicolumn{1}{c}{77.62} & \multicolumn{1}{c|}{0.00} & \multicolumn{1}{c}{15.21} & \multicolumn{1}{c}{11711} & \multicolumn{1}{c}{74.59} & \multicolumn{1}{c|}{0.00} & \multicolumn{1}{c}{32.93} & \multicolumn{1}{c}{47013} & \multicolumn{1}{c}{64.01} & \multicolumn{1}{c}{0.00} \\
		PROSER \cite{zhou2021learning}                        & \multicolumn{1}{c}{10.13} & \multicolumn{1}{c}{5452} & \multicolumn{1}{c}{76.66} & \multicolumn{1}{c|}{28.56} & \multicolumn{1}{c}{17.67} & \multicolumn{1}{c}{11057} & \multicolumn{1}{c}{73.49} & \multicolumn{1}{c|}{27.80} & \multicolumn{1}{c}{36.59} & \multicolumn{1}{c}{44403} & \multicolumn{1}{c}{62.37} & \multicolumn{1}{c}{28.45} \\
		DS \cite{miller2018dropout}                           & \multicolumn{1}{c}{8.23} & \multicolumn{1}{c}{4985} & \multicolumn{1}{c}{77.83} & \multicolumn{1}{c|}{16.50} & \multicolumn{1}{c}{13.93} & \multicolumn{1}{c}{9755} & \multicolumn{1}{c}{74.83} & \multicolumn{1}{c|}{16.28} & \multicolumn{1}{c}{31.28} & \multicolumn{1}{c}{39076} & \multicolumn{1}{c}{64.72} & \multicolumn{1}{c}{16.54} \\
		OpenDet \cite{han2022opendet}                        & \multicolumn{1}{c}{6.51} & \multicolumn{1}{c}{4308} & \multicolumn{1}{c}{77.86} & \multicolumn{1}{c|}{30.32} & \multicolumn{1}{c}{11.45} & \multicolumn{1}{c}{8600} & \multicolumn{1}{c}{75.36} & \multicolumn{1}{c|}{28.75} & \multicolumn{1}{c}{25.14} & \multicolumn{1}{c}{34382} & \multicolumn{1}{c}{64.88} & \multicolumn{1}{c}{29.70} \\ \hline
		Openset RCNN                                & \multicolumn{1}{c}{6.79} & \multicolumn{1}{c}{\textbf{3937}}  & \multicolumn{1}{c}{77.41} & \multicolumn{1}{c|}{\textbf{52.99}} & \multicolumn{1}{c}{11.60} & \multicolumn{1}{c}{\textbf{7622}} & \multicolumn{1}{c}{74.15} & \multicolumn{1}{c|}{\textbf{52.44}} & \multicolumn{1}{c}{\textbf{24.94}} & \multicolumn{1}{c}{\textbf{30770}} & \multicolumn{1}{c}{63.11} & \multicolumn{1}{c}{\textbf{53.38}} \\ \hline
	\end{tabular} }
\end{table*}

\begin{table*}[ht]
	\centering
	\setlength{\abovecaptionskip}{0.cm}
	\caption{Comparisons with other methods on GraspNet-OSOD-T1.}
	\renewcommand\arraystretch{1.5}
	\label{table:graspnet_t_1}
	\setlength{\tabcolsep}{2.6mm}
	\resizebox{\textwidth}{!}{
		\begin{tabular}{l|llll|llll|llll}
			\hline
			\multicolumn{1}{l|}{\multirow{2}{*}{Method}}   & \multicolumn{4}{c|}{GraspNet-Test-1}                                                                               & \multicolumn{4}{c|}{GraspNet-Test-2}                                                                               & \multicolumn{4}{c}{GraspNet-Test-3}                                                                               \\ \cline{2-13} 
			\multicolumn{1}{c|}{}                        & \multicolumn{1}{c}{WI$_\downarrow$}    & \multicolumn{1}{c}{AOSE$_\downarrow$}  & \multicolumn{1}{c}{mAP$_\mathcal{K}{_\uparrow}$}  & \multicolumn{1}{c|}{AP$_\mathcal{U}{_\uparrow}$}   & \multicolumn{1}{c}{WI$_\downarrow$}    & \multicolumn{1}{c}{AOSE$_\downarrow$}  & \multicolumn{1}{c}{mAP$_\mathcal{K}{_\uparrow}$}  & \multicolumn{1}{c|}{AP$_\mathcal{U}{_\uparrow}$}   & \multicolumn{1}{c}{WI$_\downarrow$}    & \multicolumn{1}{c}{AOSE$_\downarrow$}  & \multicolumn{1}{c}{mAP$_\mathcal{K}{_\uparrow}$}  & \multicolumn{1}{c}{AP$_\mathcal{U}{_\uparrow}$}   \\ \hline
			FR-CNN \cite{ren2015faster}                           & \multicolumn{1}{c}{0.09}  & \multicolumn{1}{c}{124433} & \multicolumn{1}{c}{66.95} & \multicolumn{1}{c|}{0.00} & \multicolumn{1}{c}{0.21} & \multicolumn{1}{c}{288090} & \multicolumn{1}{c}{62.57} & \multicolumn{1}{c|}{0.00} & \multicolumn{1}{c}{0.25} & \multicolumn{1}{c}{403598} & \multicolumn{1}{c}{61.53} & \multicolumn{1}{c}{0.00} \\
			PROSER \cite{zhou2021learning}                        & \multicolumn{1}{c}{0.10} & \multicolumn{1}{c}{89567} & \multicolumn{1}{c}{68.27} & \multicolumn{1}{c|}{33.70} & \multicolumn{1}{c}{0.25} & \multicolumn{1}{c}{212522} & \multicolumn{1}{c}{63.71} & \multicolumn{1}{c|}{44.33} & \multicolumn{1}{c}{0.29} & \multicolumn{1}{c}{304243} & \multicolumn{1}{c}{62.75} & \multicolumn{1}{c}{44.85} \\
			DS \cite{miller2018dropout}                           & \multicolumn{1}{c}{0.12} & \multicolumn{1}{c}{89786} & \multicolumn{1}{c}{67.60} & \multicolumn{1}{c|}{23.24} & \multicolumn{1}{c}{0.27} & \multicolumn{1}{c}{217794} & \multicolumn{1}{c}{63.06} & \multicolumn{1}{c|}{26.73} & \multicolumn{1}{c}{0.31} & \multicolumn{1}{c}{310439} & \multicolumn{1}{c}{61.89} & \multicolumn{1}{c}{26.62} \\
			OpenDet \cite{han2022opendet}                        & \multicolumn{1}{c}{0.11} & \multicolumn{1}{c}{91559} & \multicolumn{1}{c}{67.75} & \multicolumn{1}{c|}{22.50} & \multicolumn{1}{c}{0.24} & \multicolumn{1}{c}{212260} & \multicolumn{1}{c}{63.66} & \multicolumn{1}{c|}{33.30} & \multicolumn{1}{c}{0.29} & \multicolumn{1}{c}{306129} & \multicolumn{1}{c}{62.31} & \multicolumn{1}{c}{35.12} \\ \hline
			Openset RCNN                                & \multicolumn{1}{c}{\textbf{0.06}} & \multicolumn{1}{c}{\textbf{22843}}  & \multicolumn{1}{c}{\textbf{69.03}} & \multicolumn{1}{c|}{\textbf{41.72}} & \multicolumn{1}{c}{\textbf{0.19}} & \multicolumn{1}{c}{\textbf{65645}} & \multicolumn{1}{c}{\textbf{65.05}} & \multicolumn{1}{c|}{\textbf{46.20}} & \multicolumn{1}{c}{\textbf{0.22}} & \multicolumn{1}{c}{\textbf{89292}} & \multicolumn{1}{c}{\textbf{64.30}} & \multicolumn{1}{c}{\textbf{48.45}} \\ \hline
	\end{tabular} }
\end{table*}

\begin{table*}[ht]
	\centering
	\setlength{\abovecaptionskip}{0.cm}
	\caption{Comparisons with other methods on GraspNet-OSOD-T2.}
	\renewcommand\arraystretch{1.5}
	\label{table:graspnet_t_2}
	\setlength{\tabcolsep}{2.6mm}
	\resizebox{\textwidth}{!}{
		\begin{tabular}{l|llll|llll|llll}
			\hline
			\multicolumn{1}{l|}{\multirow{2}{*}{Method}}   & \multicolumn{4}{c|}{GraspNet-Test-4}                                                                               & \multicolumn{4}{c|}{GraspNet-Test-5}                                                                               & \multicolumn{4}{c}{GraspNet-Test-6}                                                                               \\ \cline{2-13} 
			\multicolumn{1}{c|}{}                        & \multicolumn{1}{c}{WI$_\downarrow$}    & \multicolumn{1}{c}{AOSE$_\downarrow$}  & \multicolumn{1}{c}{mAP$_\mathcal{K}{_\uparrow}$}  & \multicolumn{1}{c|}{AP$_\mathcal{U}{_\uparrow}$}   & \multicolumn{1}{c}{WI$_\downarrow$}    & \multicolumn{1}{c}{AOSE$_\downarrow$}  & \multicolumn{1}{c}{mAP$_\mathcal{K}{_\uparrow}$}  & \multicolumn{1}{c|}{AP$_\mathcal{U}{_\uparrow}$}   & \multicolumn{1}{c}{WI$_\downarrow$}    & \multicolumn{1}{c}{AOSE$_\downarrow$}  & \multicolumn{1}{c}{mAP$_\mathcal{K}{_\uparrow}$}  & \multicolumn{1}{c}{AP$_\mathcal{U}{_\uparrow}$}   \\ \hline
			FR-CNN \cite{ren2015faster}                           & \multicolumn{1}{c}{0.10}  & \multicolumn{1}{c}{48804} & \multicolumn{1}{c}{68.19} & \multicolumn{1}{c|}{0.00} & \multicolumn{1}{c}{0.24} & \multicolumn{1}{c}{156053} & \multicolumn{1}{c}{61.34} & \multicolumn{1}{c|}{0.00} & \multicolumn{1}{c}{0.28} & \multicolumn{1}{c}{236779} & \multicolumn{1}{c}{59.35} & \multicolumn{1}{c}{0.00} \\
			PROSER \cite{zhou2021learning}                        & \multicolumn{1}{c}{0.09} & \multicolumn{1}{c}{34878} & \multicolumn{1}{c}{69.24} & \multicolumn{1}{c|}{32.35} & \multicolumn{1}{c}{0.25} & \multicolumn{1}{c}{115422} & \multicolumn{1}{c}{62.15} & \multicolumn{1}{c|}{47.03} & \multicolumn{1}{c}{0.30} & \multicolumn{1}{c}{177889} & \multicolumn{1}{c}{60.57} & \multicolumn{1}{c}{45.02} \\
			DS \cite{miller2018dropout}                           & \multicolumn{1}{c}{0.11} & \multicolumn{1}{c}{35080} & \multicolumn{1}{c}{68.72} & \multicolumn{1}{c|}{23.95} & \multicolumn{1}{c}{0.29} & \multicolumn{1}{c}{118285} & \multicolumn{1}{c}{61.99} & \multicolumn{1}{c|}{28.87} & \multicolumn{1}{c}{0.35} & \multicolumn{1}{c}{182647} & \multicolumn{1}{c}{59.94} & \multicolumn{1}{c}{27.27} \\
			OpenDet \cite{han2022opendet}                        & \multicolumn{1}{c}{0.11} & \multicolumn{1}{c}{35560} & \multicolumn{1}{c}{69.16} & \multicolumn{1}{c|}{22.11} & \multicolumn{1}{c}{0.28} & \multicolumn{1}{c}{115177} & \multicolumn{1}{c}{62.97} & \multicolumn{1}{c|}{37.16} & \multicolumn{1}{c}{0.34} & \multicolumn{1}{c}{178976} & \multicolumn{1}{c}{60.44} & \multicolumn{1}{c}{36.55} \\ \hline
			Openset RCNN                                & \multicolumn{1}{c}{\textbf{0.05}} & \multicolumn{1}{c}{\textbf{9115}}  & \multicolumn{1}{c}{\textbf{70.35}} & \multicolumn{1}{c|}{\textbf{42.94}} & \multicolumn{1}{c}{\textbf{0.22}} & \multicolumn{1}{c}{\textbf{36221}} & \multicolumn{1}{c}{\textbf{63.74}} & \multicolumn{1}{c|}{\textbf{48.88}} & \multicolumn{1}{c}{\textbf{0.27}} & \multicolumn{1}{c}{\textbf{53047}} & \multicolumn{1}{c}{\textbf{62.46}} & \multicolumn{1}{c}{\textbf{49.34}} \\ \hline
	\end{tabular} }
	\vspace{-0.3cm}
\end{table*}
\vspace{-0.2cm}
\section{Experiments}
To validate the proposed method, we first report the open-set obejct detection performance on OpenDet benchmark \cite{han2022opendet}. As mentioned before, commonly used object detection datasets are not fully annotated and the detection performance of unknown objects can not be unbiasedly evaluated. We introduce a new benchmark by reorganizing a completely annotated dataset, GraspNet-1billion \cite{fang2020graspnet}, and the open-set obejct detection performance is also reported.
\vspace{-0.2cm}
\subsection{Metrics}
To evaluate the open-set object detection performance, we use the \textbf{Wilderness Impact} (WI) \cite{dhamija2020overlooked} to measure the degree of unknown objects misclassified to known classes: 
\begin{equation}
WI= \left( \frac{P_{\mathcal{K}}}{P_{\mathcal{K} \cup \mathcal{U}}} - 1 \right) \times 100
\end{equation}
where $P_{\mathcal{K}}$ and $P_{\mathcal{K} \cup \mathcal{U}}$ denote the precision of known classes in close-set and open-set condition, respectively. Following \cite{joseph2021towards}, we report WI under the recall level of 0.8. Besides, we also use \textbf{Absolute Open-Set Error} (AOSE) \cite{miller2018dropout} to count the number of unknown objects that get wrongly classified as any of the known class. Both WI and AOSE implicitly measure how effective the model is in handling unknown objects. Furthermore, we report the \textbf{mean Average Precision} (mAP) of known classes (mAP$_\mathcal{K}$) to measure the close-set object detection performance. Different from \cite{han2022opendet}, we use \textbf{Recall} of unknown objects (R$_\mathcal{U}$) indicating the detection ability of the unknown classes rather than Average Precision. Since \textbf{False Positive} of unknown objects can not be counted on the situation that objects in VOC and COCO are not fully annotated. Thus, WI, AOSE, and R$_\mathcal{U}$ are open-set metrics, and mAP$_\mathcal{K}$ is a close-set metric.
\vspace{-0.2cm}
\subsection{Comparison Methods}

We compare Openset RCNN with the following methods: Faster RCNN (FR-CNN) \cite{ren2015faster}, Dropout
Sampling (DS) \cite{miller2018dropout}, PROSER \cite{zhou2021learning} and OpenDet \cite{han2022opendet}. 
We do not compare with ORE \cite{joseph2021towards} because ORE relies on a validation set with annotations for the unknowns which violates the original definition of OSOD. We use the official code of OpenDet and retrain all models.

\subsection{Results on OpenDet Benchmark}

OpenDet benchmark \cite{han2022opendet} is constructed based on PASCAL VOC \cite{everingham2010pascal} and MS COCO \cite{lin2014microsoft}. The VOC07 train and VOC12 trainval splits are taken for close-set training and 20 VOC classes are regarded as known categories. To create open-set conditions, 60 non-VOC classes in COCO are set to be unknown and two settings are defined: \textbf{VOC-COCO-\{\textbf{T$_1$,T$_2$}\}}. Open-set classes are gradually increased in setting \textbf{T$_1$} and Wilderness Ratio (WR) \cite{dhamija2020overlooked} is gradually increased in setting \textbf{T$_2$}. Meanwhile, the VOC test2007 split is used to evaluate the close-set detection performance. The quantitative results on OpenDet benchmark are shown in Tab. \ref{table:opendet_t_1} and Tab. \ref{table:opendet_t_2}. Note that mAP$_\mathcal{K}$ is calculated using VOC 2012 method and the threshold used to identify known and unknown objects $T_u=0.23$. The proposed Openset RCNN outperforms other methods with a healthy margin on VOC-COCO-T1 and achieves comparable results on VOC-COCO-T2 with OpenDet. Some qualitative results are illustrated in Fig. \ref{fig:fig4}(a). More discussions about the results are in Appendix-B \cite{ours2022arxiv}.

\subsection{Results on GraspNet OSOD Benchmark}

\begin{figure*}[thpb]
	\centering
	\setlength{\abovecaptionskip}{-0.2cm}
	\includegraphics[scale=0.0805, trim={1.7cm 0 0 0}, clip]{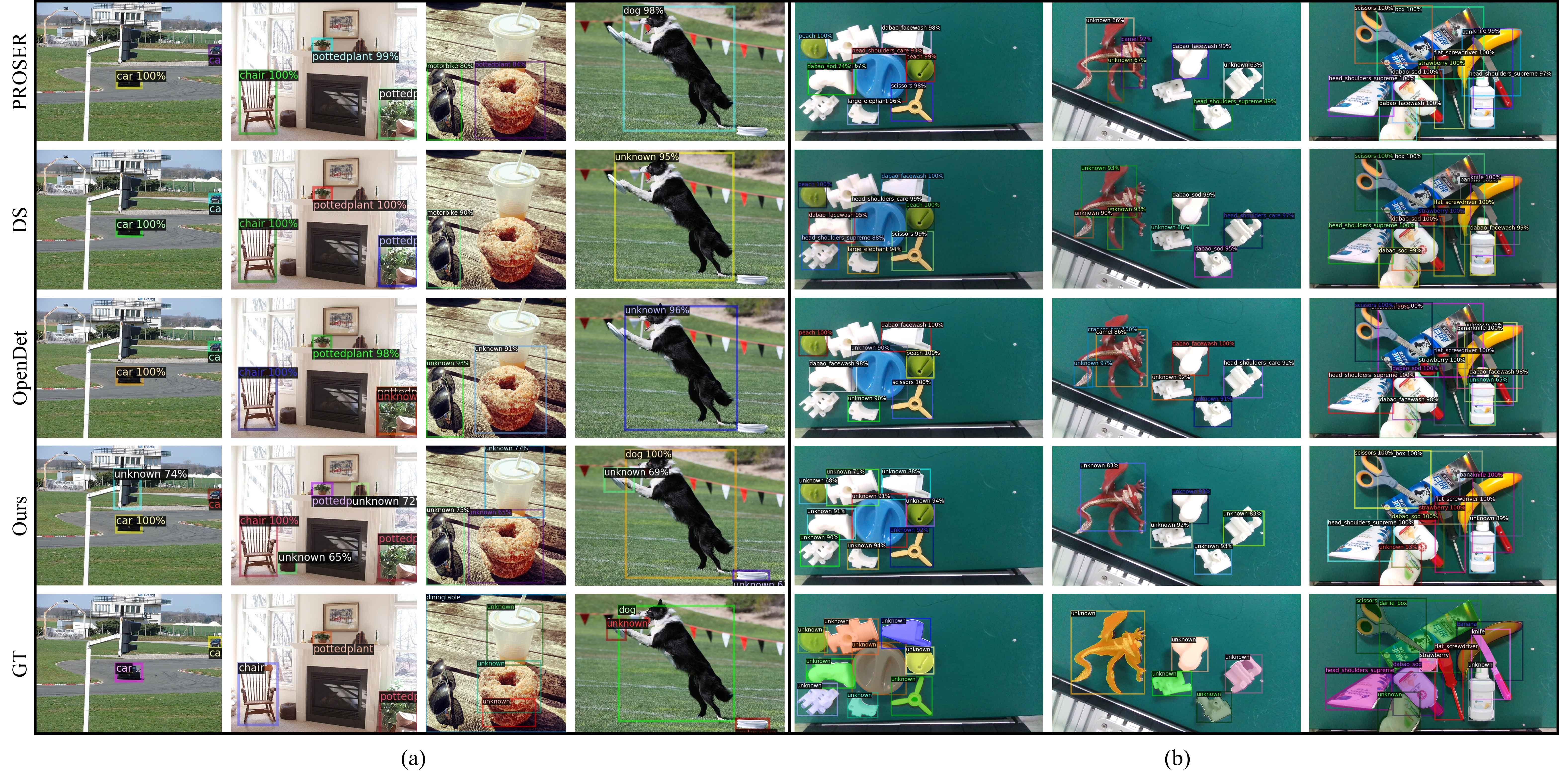}
	\caption{Qualitative comparisons between Openset RCNN and other methods on (a) OpenDet benchmark \cite{han2022opendet} and (b) GraspNet OSOD benchmark.}
	\label{fig:fig4}
	\vspace{-0.3cm}
\end{figure*}

We introduce a new benchmark by reorganizing GraspNet-1billion \cite{fang2020graspnet}, a robotic grasp pose detection dataset with complete annotation. We take 28 out of 88 classes as known categories and 9728 images are used for close-set training. Following the same way of creating open-set conditions in \cite{han2022opendet}, we also define two settings: \textbf{GraspNet-OSOD-\{\textbf{T$_1$,T$_2$}\}}. For setting T$_1$, we gradually increase open-set classes to build three joint datasets GraspNet-Test-$\{1, 2, 3\}$ containing \{23808, 31744, 38912\} images of 28 known classes and \{12, 34, 60\} unknown classes. For setting T$_2$, we gradually increase WR to construct three joint datasets GraspNet-Test-$\{4, 5, 6\}$ containing \{5120, 10240, 15360\} images with  WR=\{1, 2, 3\}. The results on GraspNet OSOD benchmark are shown in Tab. \ref{table:graspnet_t_1} and Tab. \ref{table:graspnet_t_2}. Note that mAP$_\mathcal{K}$ is calculated using COCO method and the threshold used to identify known and unknown objects $T_u=0.09$. The proposed Openset RCNN outperforms other methods remarkably on all metrics. Some qualitative results are illustrated in Fig. \ref{fig:fig4}(b). 

\begin{table}[h]
	\centering
	\setlength{\abovecaptionskip}{0.cm}
	\caption{Effect of key components on GraspNet-Test-6.}
	\renewcommand\arraystretch{1.5}
	\label{table:ablation}
	\setlength{\tabcolsep}{3.4mm}
	\begin{tabular}{cc|cccc}
		\hline
		CF-RPN               & PLN                   & \multicolumn{1}{c}{WI$_\downarrow$}    & \multicolumn{1}{c}{AOSE$_\downarrow$}  & \multicolumn{1}{c}{mAP$_\mathcal{K}{_\uparrow}$}  & \multicolumn{1}{c}{AP$_\mathcal{U}{_\uparrow}$} \\ \hline
		\multicolumn{2}{c|}{baseline}  & 0.28 & 254027  & 61.15  & 0.00  \\
		$\checkmark$  &                & 0.29 & 236594  & 61.24  & 0.00  \\
		              & $\checkmark$   & 0.39 & 69339   & 61.97  & 37.30 \\
		$\checkmark$ & $\checkmark$    & \textbf{0.27} & \textbf{53047} & \textbf{62.46} & \textbf{49.34} \\ \hline
	\end{tabular}
	\vspace{-0.2cm}
\end{table}

\subsection{Comparisons with OW-DETR}

We make evaluations using the protocol proposed by ORE which is the same as OW-DETR \cite{gupta2022ow}. The protocol was designed for open-world object detection including four tasks to evaluate the performance of open-set object detection as well as incremental learning. Since incremental learning is not within the scope of our work, we only evaluate the proposed method on task 1 and make comparisons with OW-DETR. The results are reported in Tab. \ref{table:comparison_owdetr} showing the superiority of our Openset RCNN.

\begin{table}[thpb]
	\centering
	\setlength{\abovecaptionskip}{0.cm}
	\caption{Comparisons between Openset RCNN and OW-DETR.}
	\renewcommand\arraystretch{1.5}
	\label{table:comparison_owdetr}
	\setlength{\tabcolsep}{3.9mm}{
		\begin{tabular}{l|llll}
			\hline
			\multicolumn{1}{l|}{\multirow{2}{*}{Method}}   & \multicolumn{4}{c}{OWOD-Task-1} \\ \cline{2-5} 
			& WI$_\downarrow$ & AOSE$_\downarrow$ & mAP$_\mathcal{K}{_\uparrow}$ & R$_\mathcal{U}{_\uparrow}$    \\ \hline
			OW-DETR        & 5.71  &  10240  & 59.21  & 7.51  \\
			Openset RCNN   & \textbf{4.62}  &  \textbf{5707}   & \textbf{59.31}  & \textbf{22.22} \\ \hline
	\end{tabular} }
	\vspace{-0.5cm}
\end{table}

\subsection{Ablation Studies}

We conduct ablation experiments on GraspNet-Test-6 to analyze the effect of our main components. Our baseline is based on Faster R-CNN \cite{ren2015faster} consisting of a ResNet-50 backbone, standard Region Proposal Network and R-CNN in which box regressor is set to class-agnostic. As shown in Tab. \ref{table:ablation}, the proposed two modules, CF-RPN and PLN, reduce AOSE substantially. Because unknown objects are modeled using the complementary region of known categories in latent space, AP$_\mathcal{U}$ of the PLN equipped baseline model (37.30) outperforms DS \cite{miller2018dropout} and OpenDet \cite{han2022opendet} (27.27 and 36.55 in Tab. \ref{table:graspnet_t_2}). And the combination of CF-RPN and PLN further improves the performance which verifies the effectiveness of the proposed components.

\subsection{Application in Robotic Rearrangement Tasks}

\begin{figure}[h]
	\centering
	\setlength{\abovecaptionskip}{0.cm}
	\includegraphics[scale=0.09]{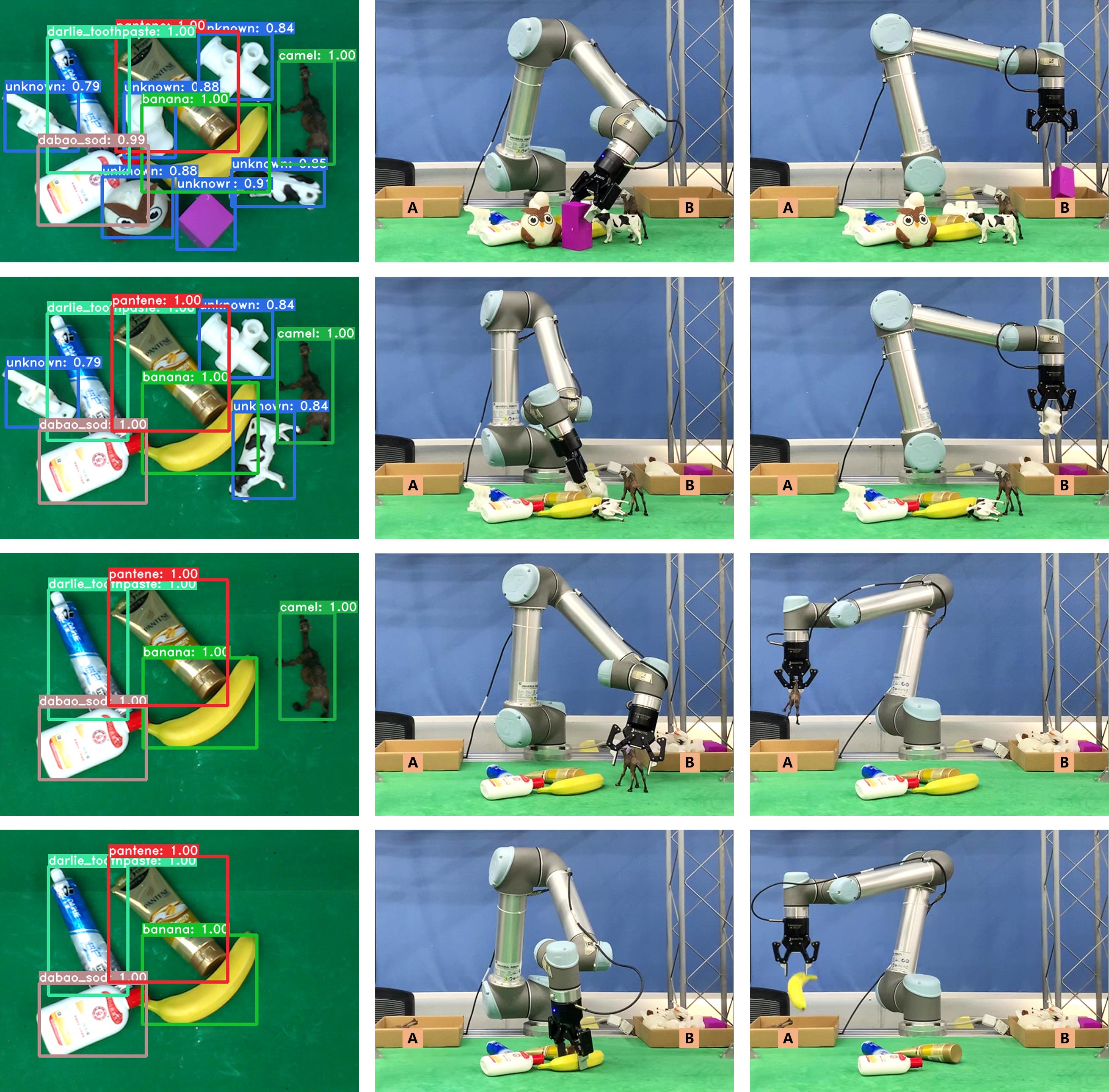}
	\caption{Visualization of robotic rearrangement using the proposed detection method and 6-DOF GraspNet \cite{fang2020graspnet}.}
	%\vspace{-0.3cm}
	\label{fig:fig5}
	\vspace{-0.6cm}
\end{figure}

\begin{figure}[h]
	\centering
	\setlength{\abovecaptionskip}{-0.1cm}
	\includegraphics[scale=0.40]{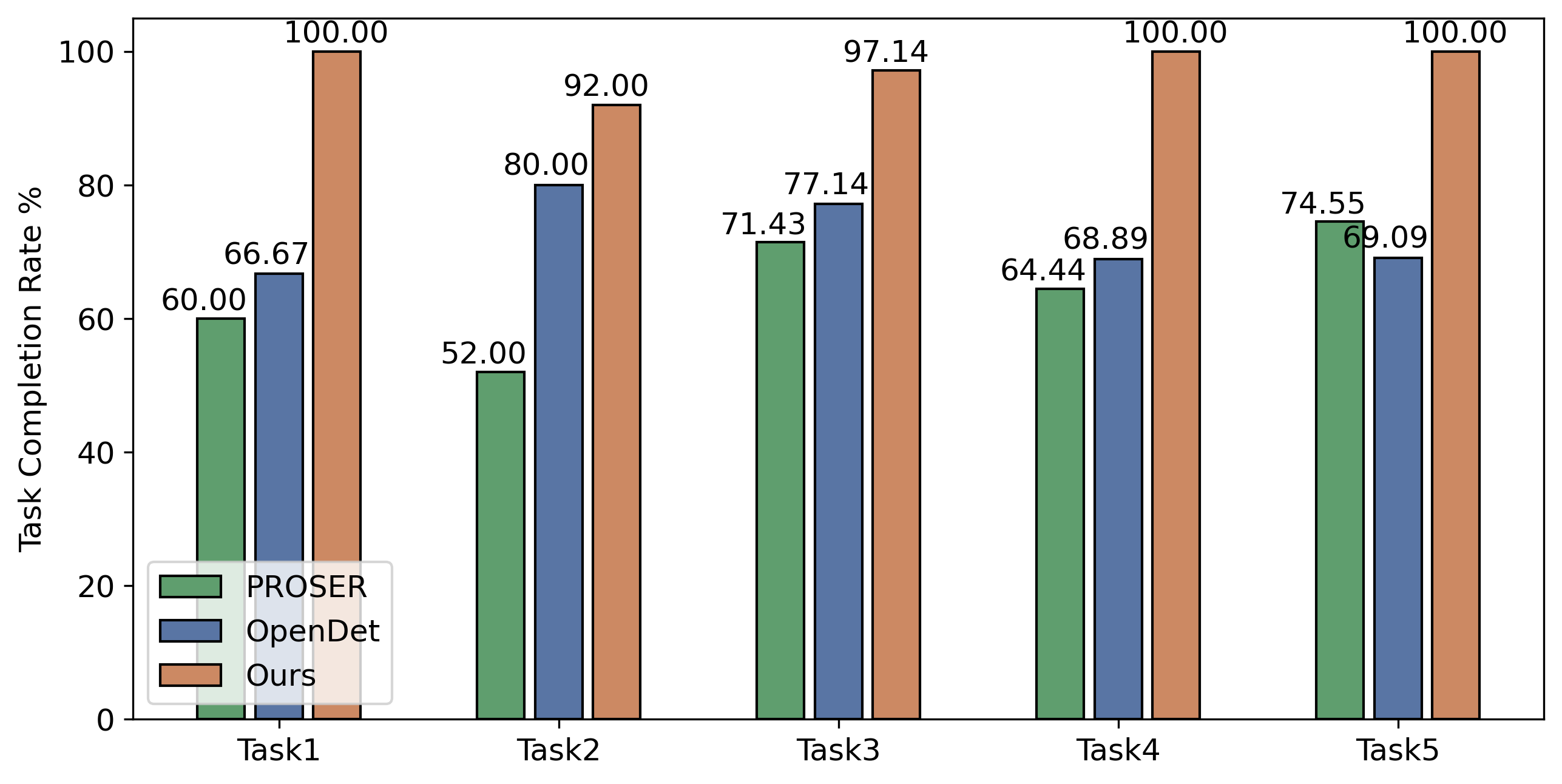}
	\caption{The average task completion rate of 5 robotic rearrangement tasks using PROSER \cite{zhou2021learning}, OpenDet \cite{han2022opendet} and the proposed Openset RCNN.}
	\label{fig:fig6}
	\vspace{-0.3cm}
\end{figure}

We use the proposed method to demonstrate robotic object rearrangement in a cluttered environment using a UR5 collaborative robot with a Robotiq gripper and an external RGB-D camera. The task is to collect objects from a table and put known objects in bin A and unknown objects in bin B. Objects are detected using the proposed method and the point cloud of the scene is fed to GraspNet \cite{fang2020graspnet} to generate 6-DOF candidate grasps. The grasp that has the maximum score and locate within the detected bounding box is chosen to execute. Fig. \ref{fig:fig5}  shows the open-set detection results at different stages of the task and also the execution of the robot grasps. Further, we compare the performance of PROSER \cite{zhou2021learning}, OpenDet \cite{han2022opendet} and the proposed Openset RCNN in five robotic rearrangement tasks containing \{3, 5, 7, 9, 11\} objects respectively. Each task is conducted 5 times. The average task completion rate is reported in Fig. \ref{fig:fig6} and the proposed method outperforms the others. In detail, we investigate the number of objects wrongly detected and the results are shown in Fig. \ref{fig:fig7}. More experimental results of each task can be found in Appendix-D \cite{ours2022arxiv}.

\begin{figure}[h]
	\centering
	\setlength{\abovecaptionskip}{-0.1cm}
	\includegraphics[scale=0.403]{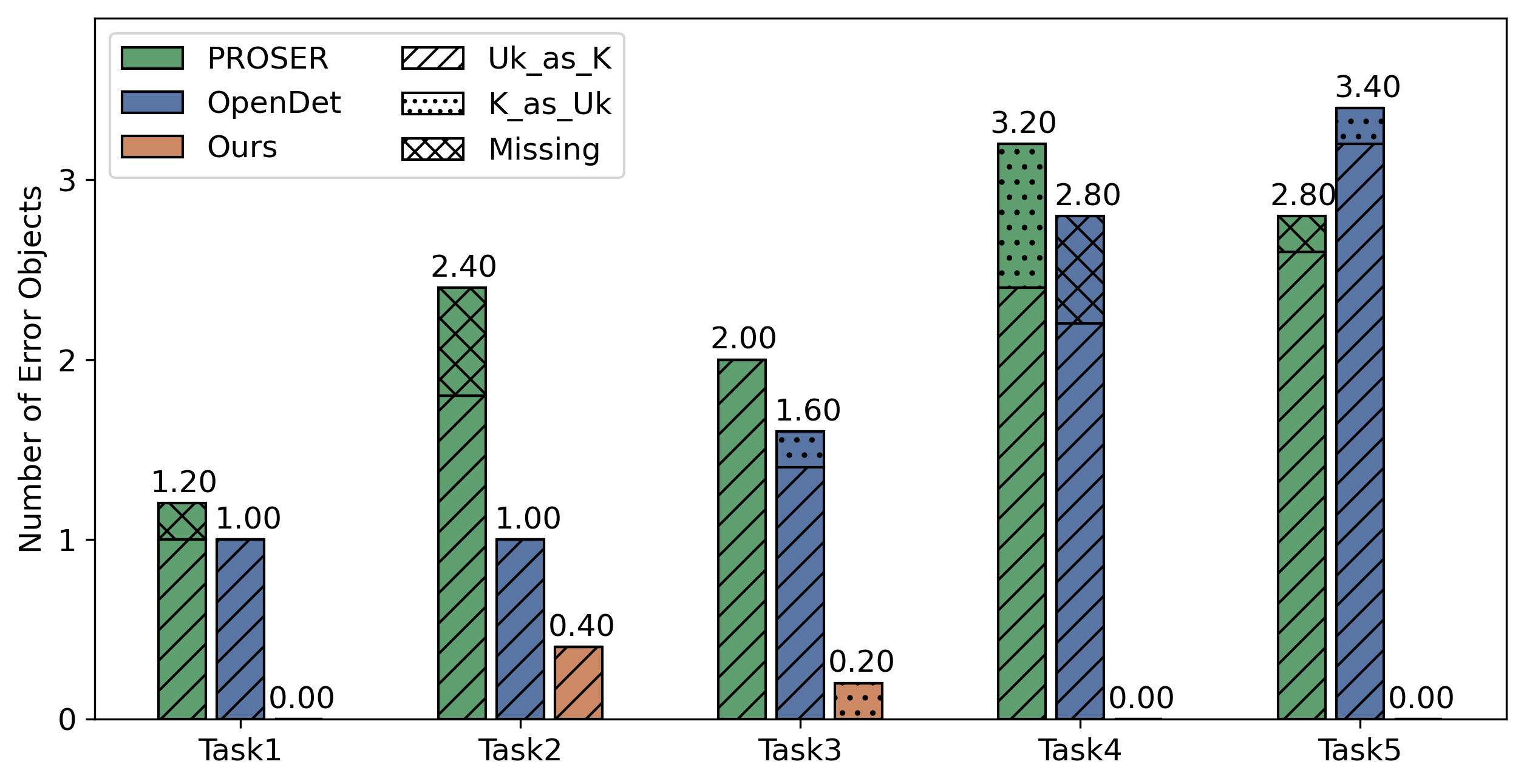}
	\caption{The number of wrongly detected objects in 5 robotic rearrangement tasks using PROSER \cite{zhou2021learning}, OpenDet \cite{han2022opendet} and the proposed Openset RCNN. ``Uk\_as\_K'' means unknown objects are detected as one of the known categories. ``K\_as\_Uk'' means known objects are detected as unknown and ``Missing'' means objects are detected as background.}
	\vspace{-0.5cm}
	\label{fig:fig7}
\end{figure}

\section{Conclusions}

In this work, we investigate the challenging open-set object detection and propose a novel method, Openset RCNN, with enhanced generalization ability for unknown object proposals as well as open-set classification. We also build an OSOD benchmark by reorganizing GraspNet-1billion dataset and conduct extensive experiments to demonstrate the effectiveness of our method. Finally, we show that our method can be applied to the robotic rearrangement task in cluttered environments supporting the robot to manipulate known and unknwon objects differently. More analysis and experiments can be accessed at Appendix \cite{ours2022arxiv}.

%\addtolength{\textheight}{-12cm}   % This command serves to balance the column lengths
                                  % on the last page of the document manually. It shortens
                                  % the textheight of the last page by a suitable amount.
                                  % This command does not take effect until the next page
                                  % so it should come on the page before the last. Make
                                  % sure that you do not shorten the textheight too much.

%%%%%%%%%%%%%%%%%%%%%%%%%%%%%%%%%%%%%%%%%%%%%%%%%%%%%%%%%%%%%%%%%%%%%%%%%%%%%%%%

%%%%%%%%%%%%%%%%%%%%%%%%%%%%%%%%%%%%%%%%%%%%%%%%%%%%%%%%%%%%%%%%%%%%%%%%%%%%%%%%

%%%%%%%%%%%%%%%%%%%%%%%%%%%%%%%%%%%%%%%%%%%%%%%%%%%%%%%%%%%%%%%%%%%%%%%%%%%%%%%%

%%%%%%%%%%%%%%%%%%%%%%%%%%%%%%%%%%%%%%%%%%%%%%%%%%%%%%%%%%%%%%%%%%%%%%%%%%%%%%%%

\bibliographystyle{IEEEtran} % use IEEEtran.bst style
\bibliography{IEEEabrv,./manuscript}

\end{document}